\documentclass[aps,prl,reprint,groupedaddress]{revtex4-1}

\usepackage{multirow}
\usepackage{amsmath}
\usepackage{amssymb}
\usepackage{amstext}
\usepackage{epsfig}
\usepackage{graphicx}
\usepackage{bm}
\usepackage{eufrak}
\usepackage{subfigure}
\usepackage{appendix}
\usepackage{color}

\def\yhat{\hat{y}}

\begin{document}


\title{Model-free prediction of noisy chaotic time series by Deep Learning}


\author{Kyongmin Yeo}
\email[Corresponding author: ]{kyeo@us.ibm.com}
\affiliation{IBM T.J. Watson Research Center, Yorktown Heights, NY}


\date{\today}

\begin{abstract}
We present a deep neural network for a model-free prediction of a chaotic dynamical system from noisy observations. The proposed deep learning model aims to predict the conditional probability distribution of a state variable. The Long Short-Term Memory network (LSTM) is employed to model the nonlinear dynamics and a softmax layer is used to approximate a probability distribution. The LSTM model is trained by minimizing a regularized cross-entropy function. The LSTM model is validated against delay-time chaotic dynamical systems, Mackey-Glass and Ikeda equations. It is shown that the present LSTM makes a good prediction of the nonlinear dynamics by effectively filtering out the noise. It is found that the prediction uncertainty of a multiple-step forecast of the LSTM model is not a monotonic function of time; the predicted standard deviation may increase or decrease dynamically in time.
\end{abstract}


\maketitle


Data-driven reconstruction of a dynamical system has been of great interest due to its direct relevance to many applications in physics, biology, and engineering. There has been a significant progress in the data-driven modeling of a nonlinear dynamical system when the observation is noise-free or a priori information on the system is available\cite{Julier97,Hamilton15,Raissi17,Wang16}. However, there are only a very limited number of methods for a chaotic, nonlinear dynamical system, when the data is corrupted by a noise and the underlying system is completely unknown\cite{Hamilton16}. 

Here, we employ a deep-learning model for data-driven simulations of noisy nonlinear dynamical systems. Recently, deep learning has attracted great attention because of its strong capability in discovering complex structures in data \cite{LeCun15}. Although deep learning has been shown to outperform the conventional statistical methods for the data mining problems, e.g., speech recognition, image classification/identification, there are only a small number of studies on the behaviors of deep learning for complex dynamical systems \cite{Trischler16,Rivkind17}. Considering its strength in learning a nonlinear manifold of the data and the de-noising capability\cite{Bengio13}, deep learning has a potential to provide an effective tool for data-driven reconstruction of noisy dynamical system. 

In this study, we consider the following delay-time dynamical system,
\begin{equation}
\frac{d y}{dt} = f(y(t),y(t-\tau)),
\end{equation}
in which $f(\cdot,\cdot)$ is a nonlinear function and $\tau$ is a time-delay parameter. Further, we assume that the underlying dynamical system, or the ground truth, is not observable. We observe only a discrete, noisy time series,
\begin{equation}
\yhat_t = y(t)+\epsilon,
\end{equation}
in which $\epsilon$ is a white Gaussian noise, $\epsilon \sim \mathcal{N}(0,\sigma^2)$. The sampling interval of the time series is denoted by $\delta t$, i.e., $\yhat_{t+1} = y(t+\delta t)+\epsilon$. Clearly, $\yhat_t$ is a random variable; $\yhat_t \sim \mathcal{N}(y(t),\sigma^2)$. Here, we are interested in the forecast of the probability distribution of $\yhat$ conditioned on the noisy observations, i.e., $p(\yhat_{t+n}|\widehat{\bm{Y}}_{0:t})$ for $n\ge1$, where $\widehat{\bm{Y}}_{0:t} = (\yhat_0,\cdots,\yhat_t)$ is the noisy observation up to the current time, $t$.

To model the dynamical system, we employ the Long Short-Term Memory network (LSTM), which is capable of memorizing a long delay-time structure in data\cite{Hochreiter97,Gers00}. The following LSTM architecture is considered;
\begin{itemize}
  \item Input network:
  \begin{align}
  \bm{z} = \mathcal{W}^{N_c,N_c} \left(\varphi_T \circ \mathcal{W}^{N_c,1} (\yhat_t)+\bm{h}_{t-1} \right). \label{eqn:LSTM_1}
  \end{align}
  \item LSTM gating functions:
  \begin{align}
  &\bm{G}_i = \bm{\varphi}_S \circ \mathcal{W}^{N_c,N_c}( \bm{z} ), ~\text{for}~i=1,2,3, \label{eqn:LSTM_2}\\
  &\bm{G}_4 = \bm{\varphi}_T \circ \mathcal{W}^{N_c,N_c}( \bm{z} ). \label{eqn:LSTM_3}
  \end{align} 
  \item LSTM internal state:
  \begin{align}
  \bm{s}_{t} = \left( \bm{1} - \bm{G}_1 \right) \odot \bm{s}_{t-1} + \bm{G}_2 \odot \bm{G}_4. \label{eqn:LSTM_4}
  \end{align} 
  \item Output network:
  \begin{align}
  &\bm{h}_{t} = \bm{G}_3 \odot \bm{s}_t, \label{eqn:LSTM_5}\\
  &\bm{o} = \mathcal{W}^{N_o,N_c}(\varphi_T \circ \mathcal{W}^{N_c,N_c}( \varphi_{SP} \circ \mathcal{W}^{N_c,N_c} (\bm{h}_t))), \label{eqn:LSTM_6}\\
  &\bm{P} = \varphi_{SM}(\bm{o}). \label{eqn:LSTM_7}
  \end{align}
\end{itemize}
Here, $N_c$ is the dimension of the LSTM and $N_o$ is the length of the output vector. $\varphi_T$, $\varphi_S$, and $\varphi_{SP}$ represent an element-wise operation of the hyperbolic tangent, Sigmoid, and softplus functions\cite{GoodfellowBengio16}, respectively, and $\bm{a}\odot\bm{b}$ denotes an element-wise multiplication. $\mathcal{W}^{a,b}$ is a linear transformation operator;
\[
\mathcal{W}^{a,b}(\bm{x}) = \bm{W}\bm{x} + \bm{c},
\]
in which $\bm{W}\in\mathbb{R}^{a \times b}$ is a weight matrix and $\bm{c}\in\mathbb{R}^a$ is a bias vector. 
The last layer of the proposed LSTM is the softmax function, which is defined as
\begin{equation}
P_i = \frac{exp(o_i)}{\sum_{j=1}^{N_o} exp(o_j)},~~\text{for}~~i=1,\cdots,N_o.
\end{equation}
The output vector of the LSTM, $\bm{P}$, defines a discrete probability distribution, because $P_i \ge 0$ and $\sum_i P_i = 1$. 

To assign a probability distribution to $\bm{P}$, we assume that $\bm{P}$ is a discretization of $p(\yhat_{t+1}|\widehat{\bm{Y}}_{0:t})$;
\begin{equation} \label{eqn:discret}
P_i = \int_{\alpha_i}^{\alpha_{i+1}} p(\yhat_{t+1}|\widehat{\bm{Y}}_{0:t}) d\yhat_{t+1}.
\end{equation}
Here, $\bm{\alpha} = (\alpha_1,\cdots\alpha_{N_o+1})$ is a set of ordered real numbers, which represents the boundaries of a discretization interval. In other words, $P_i$ corresponds to probability of $\yhat_{t+1}\in(\alpha_i,\alpha_{i+1})$, or, $P_i = p(i|\bm{\theta})$, where $\bm{\theta}$ is the parameters of the LSTM. For simplicity, we omit the dependence on the past trajectory, $\widehat{\bm{Y}}_{0:t}$, in the notation for now. The parameters, $\bm{\theta}$, are the weights ($\bm{W}$) and biases ($\bm{c}$) of $\mathcal{W}$'s in (\ref{eqn:LSTM_1}--\ref{eqn:LSTM_7}).

Usually, $\bm{\theta}$ is estimated from a minimum negative log-likelihood (NLL) method. After the discretization (\ref{eqn:discret}), the problem is converted to a classification task, of which NLL is the generalized Bernoulli distribution\cite{Bishop06};
\begin{equation} \label{eqn:LLH}
- \log p(\bm{l}|\bm{\theta}) = - \sum_{n=1}^N \sum_{i=1}^{N_o} {\delta_{l_n i}} \log P^n_i(\bm{\theta}).
\end{equation} 
Here, $N$ is the total number of the data, $l_n$ is the index of the interval ($\bm{\alpha}$) for $n$-th data, e.g., $\alpha_{l_n} < \yhat^n_{t+1} < \alpha_{l_n+1}$, $\bm{l} = (l_1,\cdots,l_N)$, $\delta_{ij}$ is the Kronecker delta, and $\bm{P}^n$ denotes the LSTM output for the $n$-th data. This type of minimum NLL, called the cross-entropy (CE) minimization, is one of the most widely used method in training a deep neural network. 

Note that this type of CE, (\ref{eqn:LLH}), does not consider smoothness of $\bm{P}$. For example, since we are interested in approximating a smooth probability distribution, $p(\yhat_{t+1})$, we expect that $P_i$ is close to $P_{i\pm1}$. However, such proximity structure is not considered in (\ref{eqn:LLH}). To impose a smoothness condition, we propose a regularized CE;
\begin{equation}\label{eqn:RCE}
\mathcal{L}(\bm{\theta}) = \sum_{n=1}^N \left\{ \sum_{i=1}^{N_o} -\delta_{l_n i} \log P^n_i + L_i(\bm{P}^n) \right\}.
\end{equation}
Here, $L_i$ is an $l_2$-Laplacian operator;
\[
L_i(\bm{P}) = \lambda  \left(P_{i-1}-2P_i+P_{i+1}\right)^2,
\] 
where $\lambda$ is a penalty parameter. The regularization imposes a smoothness condition by penalizing local maxima or minima. The regularized CE is solved by the standard Back-Propagation Through Time (BPTT)\cite{GoodfellowBengio16}.

Once the LSTM is trained, the predictive distribution of $\yhat$ is simply, $\bm{P} = \bm{\Psi}(\yhat_t, \bm{s}_{t-1})$, in which $\bm{\Psi}$ represents the LSTM in (\ref{eqn:LSTM_1}--\ref{eqn:LSTM_7}). Then, the moments of $\bm{P}$ can be easily calculated. For example, the expectation is
\begin{equation} \label{eqn:expect}
E[\yhat_{t+1}|\yhat_t,\bm{s}_{t-1}] = \sum_{i=1}^{N_o} \alpha_{i+1/2} P_i.\\
\end{equation}
Here, $\alpha_{i+1/2} = 0.5(\alpha_i + \alpha_{i+1})$. Note that in (\ref{eqn:expect}) the dependence on $\widehat{\bm{Y}}_{0:t}$ is replaced by ($\yhat_t,\bm{s}_{t-1})$, because LSTM provides a state-space model in which $\yhat_{t+1}$ becomes conditionally independent from $\widehat{\bm{Y}}_{0:t-1}$, given $\bm{s}_{t-1}$\cite{Bishop06}. The standard deviation (STD) or higher order moments can be calculated in the same way.

A multiple-step forecast is made by a Monte Carlo method as follows;
\begin{enumerate}
  \item Perform a sequential update of LSTM up to the last observation; $\widehat{\bm{Y}}_{0:t}$. 
  \[
\bm{P}_{i+1} = \bm{\Psi}(\yhat_i,\bm{s}_{i-1})~~\text{for}~i = 1,\cdots,t.
  \]
  \item For $N_s$ Monte Carlo samples, make $N_s$ replicas of the internal state, $\bm{s}^1_t=\cdots=\bm{s}^{N_s}_t=\bm{s}_t$, and the LSTM output, $\bm{P}^1_{t+1} = \cdots = \bm{P}^{N_s}_{t+1} = \bm{P}_{t+1}$.
  \item Sample the prediction, $\tilde{y}^i_{t+1}$, from $\bm{P}^i_{t+1}$
  \[
  \tilde{y}^i_{t+1} \sim \bm{P}^i_{t+1}, ~~\text{for}~~i = 1,\cdots,N_s.
  \]
  \item Update the probability distribution;
  \[
  \bm{P}^i_{t+2} = \bm{\Psi}(\tilde{y}^i_{t+1},\bm{s}^i_t)~~\text{for}~i = 1,\cdots,N_s.
  \]
  \item Repeat steps 3 -- 4 for a forecast horizon. The predictive probability distribution, $p(\yhat_{t+n}|\widehat{\bm{Y}}_{0:t})$, can be obtained by a density estimation from the sample trajectories, $\bm{\tilde{y}}_{t+n}$.
\end{enumerate}

The LSTM is tested against noisy observations of Mackey-Glass and Ikeda equations. The dimension of LSTM is $N_c=128$. The regularized CE is minimized by using ADAM\cite{Kingma15} with the learning rate of $10^{-3}$ and the mini-batch size of 20. In the training, BPTT is performed for 100 time steps. The training data is a time series of length $T=1.6\times10^5 \delta t$ and another time series of length $T = 2\times10^3 \delta t$ is used for the validation. The initial state of LSTM is set to zero, i.e., $\bm{s}_0 = \bm{h}_0 = \bm{0}$. Here, the LSTM is used to estimate, $p(d\yhat_{t+1}|\widehat{\bm{Y}}_{0:t})$, in which $d\yhat_{t+1} = \yhat_{t+1}-\yhat_t$. It is trivial to recover $p(\yhat_{t+1}|\widehat{\bm{Y}}_{0:t})$ from $p(d\yhat_{t+1}|\widehat{\bm{Y}}_{0:t})$. A uniform grid is used for the discretization, i.e., $\alpha_{i+1}-\alpha_i = \Delta_\alpha$ for all $i$.

First, we consider the Mackey-Glass equation \cite{Mackey77},
\begin{equation} \label{eqn:MG}
\frac{d y}{dt} = \frac{\alpha y(t-\tau)}{1+y^\beta(t-\tau)} - \gamma y(t).
\end{equation}
The parameters are $\alpha = 0.2$, $\beta = 10$, $\gamma = 0.1$, and $\tau = 17$\cite{Sprott}. Equation (\ref{eqn:MG}) is solved by a third-order Adam-Bashforth method with the time step size of 0.02. The noisy observations are generated with a sampling interval, $\delta t = 1$, and the noise ratio, $\sigma = 0.3 sd[y]$, in which $sd[y]$ is the standard deviation of $y(t)$. The discretization interval is $\Delta_\alpha = 0.02 sd[y]$ and the penalty parameter $\lambda = 200$ is used.

\begin{figure}
  \includegraphics[width=0.23\textwidth]{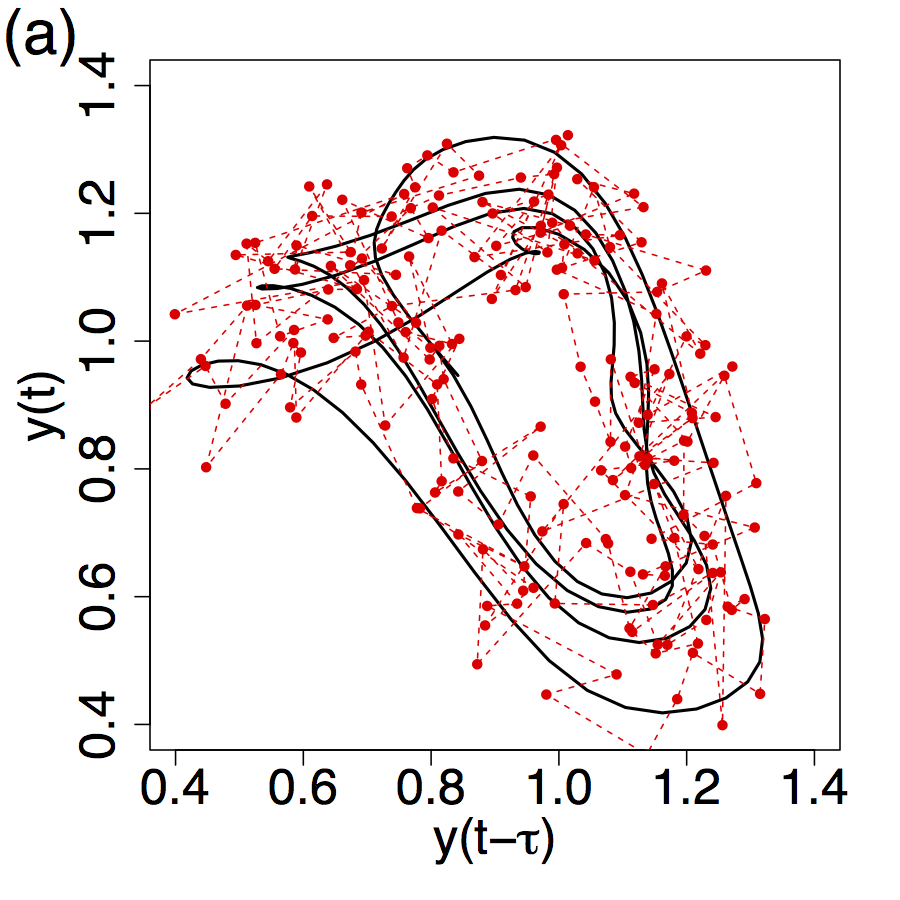}
  \includegraphics[width=0.23\textwidth]{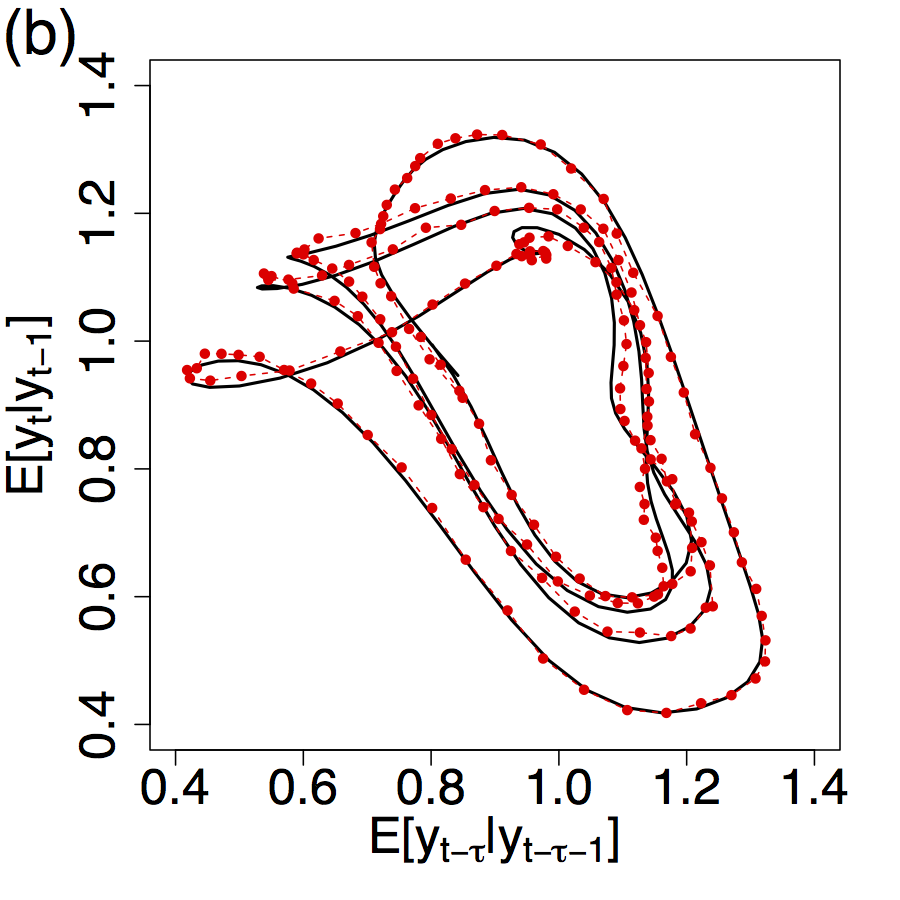}\\
  \includegraphics[width=0.23\textwidth]{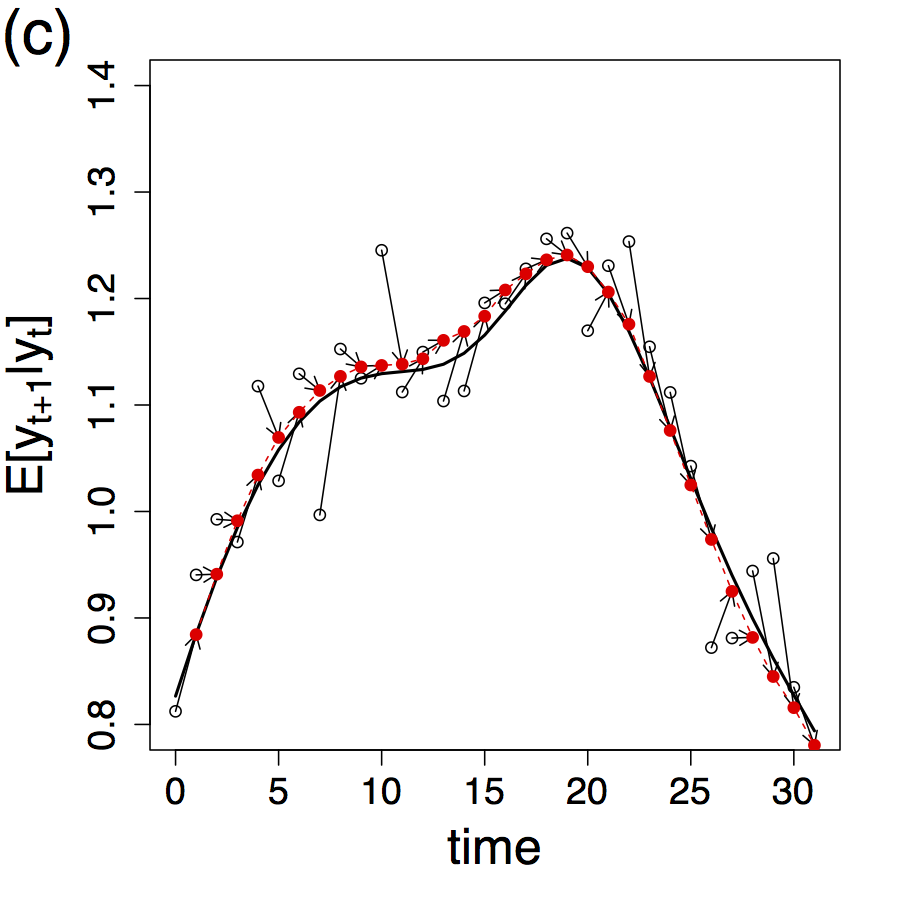}
  \includegraphics[width=0.23\textwidth]{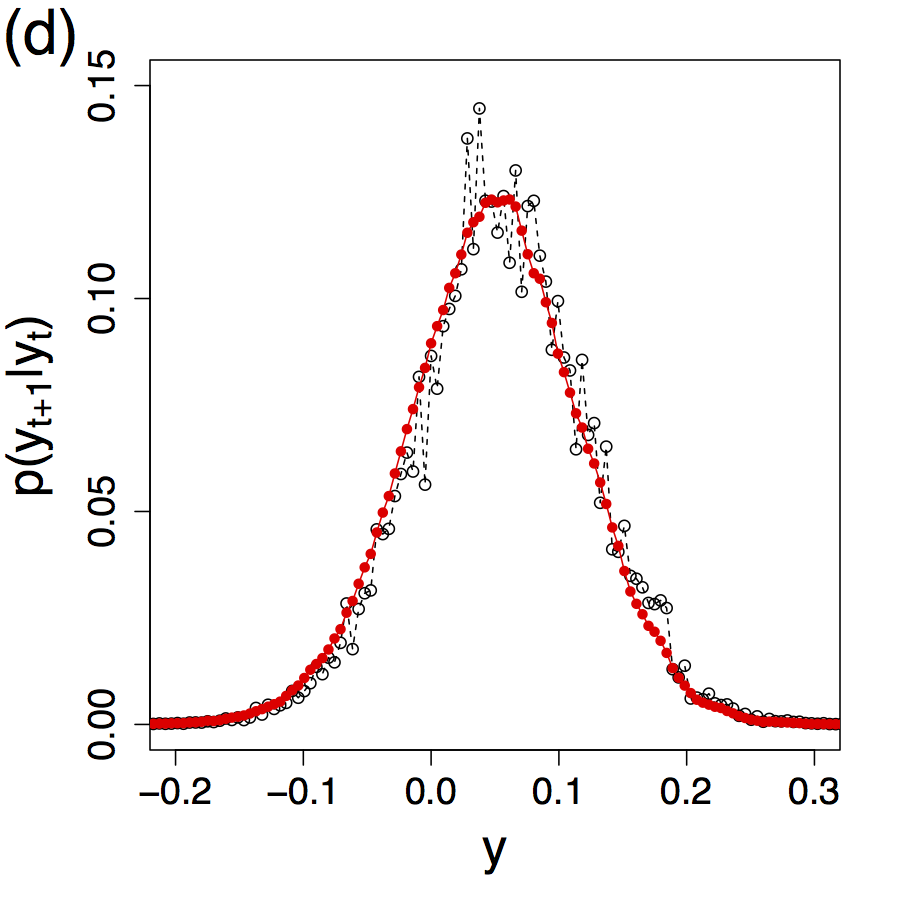}
  \caption{ (a) and (b) show the trajectories in a delay-time coordinate. The solid line denotes the ground truth and the (red) circles are (a) the noisy observations and (b) the LSTM predictions. (c) shows the next-step prediction by LSTM. The solid line ($\frac{~~~}{~~~}$) denotes the ground truth and the hollow ($\circ$) and the solid ({\color{red}$\bullet$}) circles are the noisy observation and the LSTM prediction, respectively. (d)  shows the predictive probability distributions from ($\circ$) the standard CE and ({\color{red}$\bullet$}) the regularized CE.}\label{fig:MG_pred}
\end{figure}

Figure \ref{fig:MG_pred} (a) and (b) show the phase portraits of $\yhat$ and the next-step predictions of the LSTM. The LSTM prediction is made from $\yhat$ in Fig. \ref{fig:MG_pred} (a). While the observation is so noisy that it is difficult to find a correlation between $y(t)$ and $\yhat$, the LSTM prediction approximates the phase portrait of $y(t)$ very well, suggesting that the present LSTM is able to filter out the noise and reliably recover the original nonlinear dynamics. Note that the LSTM model has never seen $y(t)$. The root mean-square error (RMSE) between $E[\yhat_{t+1}|\yhat_t,\bm{s}_{t-1}] $ and $y(t+\delta t)$ is only about 22\% of the noise, i.e., $\langle [E[\yhat_{t+1}|\yhat_t,\bm{s}_{t-1}] - y(t+\delta t)]^2 \rangle^{1/2} = 0.215 sd[\epsilon]$. Hereafter, the obvious dependence on $\bm{s}_{t-1}$ is omitted for simplicity.

Figure \ref{fig:MG_pred} (c) displays how the next-step prediction is made.  At every time step, the noisy observation is supplied to the LSTM and the prediction is made as $E[\yhat_{t+1}|\yhat_t] = \yhat_{t} + E[d \yhat_{t+1}|\yhat_t]$. In order to make a good prediction, the LSTM needs to know not only the underlying dynamics, but also how far $\yhat_t$ is from $y(t)$.

A snapshot of the predicted probability distribution is shown in figure \ref{fig:MG_pred} (d). Clearly, the regularized CE results in a smoother distribution. The ratio of the estimated STD to the noise is close to one; $sd[\yhat_{t+1}|\yhat_t] / sd[\epsilon] = 1.036$.

\begin{figure}
  \includegraphics[width=0.49\textwidth]{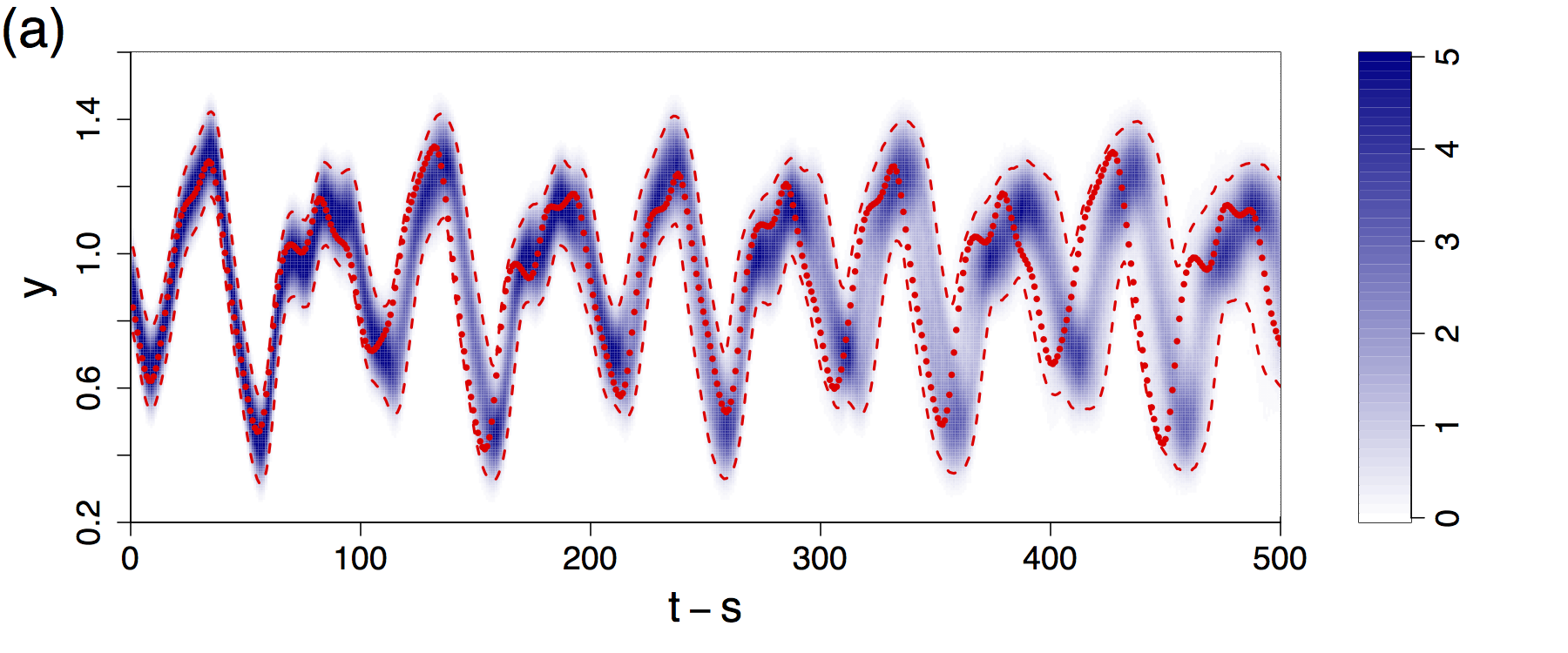}\\
  \includegraphics[width=0.49\textwidth]{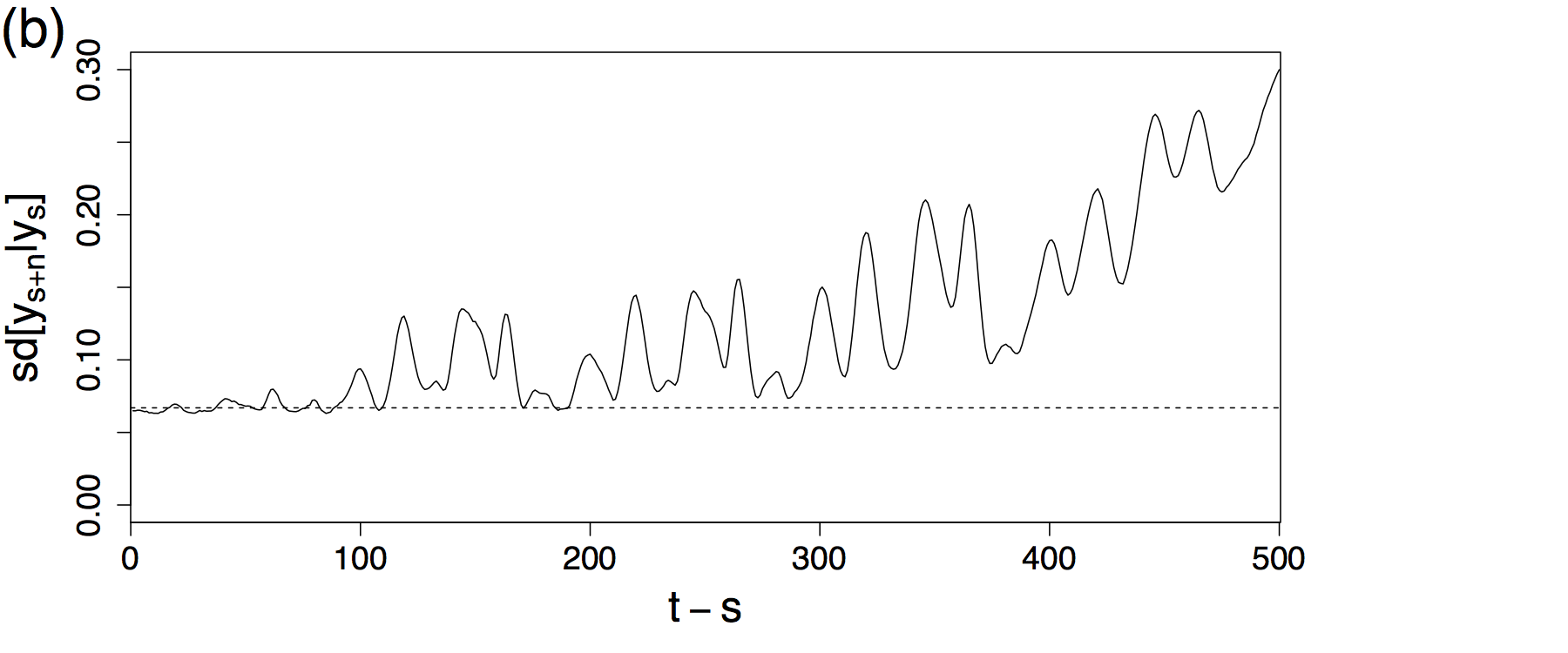}
  \caption{ (a) A 500-step forecast of the Mackey-Glass equation. The contours represent the probability distribution conditioned on the last observation; $p(\yhat_{s+n}|\yhat_s)$. The dashed lines indicate the 95\% confidence interval and the solid circles ({\color{red}$\bullet$}) are the ground truth. (b) shows the predicted STD in time. The dashed line indicates the noise level.}\label{fig:MG_dist}
\end{figure}

Figure \ref{fig:MG_dist} shows a multiple-step forecast with $N_s = 2\times10^4$. The noisy observations are used for the first 100 time steps to develop $\bm{s}_t$ from the zero initial condition. Then, a 500-step forecast is made for $t = 101 \sim 600 \delta t$. In figure \ref{fig:MG_dist} (a), it is observed that, for the initial 90 steps, $y(t)$ lies on a high probability region, then, for $t > 100$, $y(t)$ starts to deviate from the high probability region. But, it is shown that even for the 500-step forecast, $y(t)$ lies in the 95\% confidence interval.

Making a multiple-step forecast corresponds to propagating uncertainty in time. One of the measures of the uncertainty is STD. In figure \ref{fig:MG_dist} (b), the predicted STD is shown as a function of time. For conventioal linear time series models, typically the uncertainty is a non-decreasing function of time. But, for the LSTM, it is shown that STD may dynamically change in time. For the first 90 steps, the estimated STD stays that of the noise level, then starts to grow for $t > 100$. The general trend of STD is an increasing function of time, because the underlying dynamical system is chaotic. But, locally the uncertainty may increase or decrease depending on the dynamics.

\begin{figure}
  \includegraphics[width=0.23\textwidth]{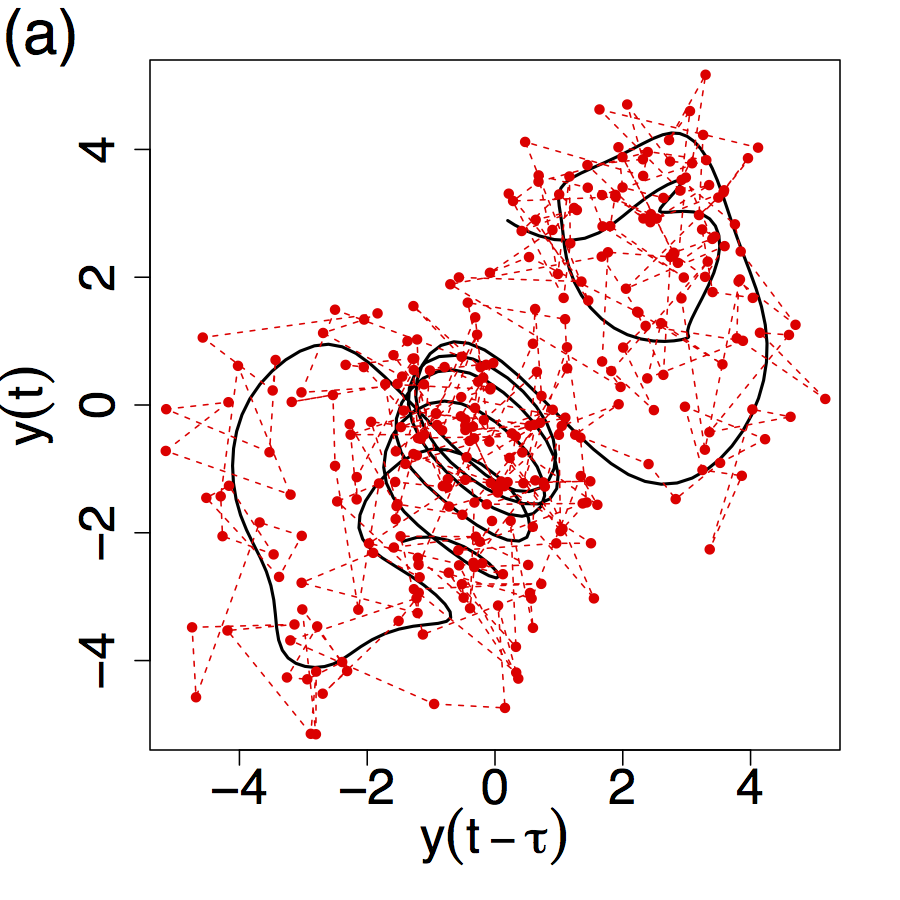}
  \includegraphics[width=0.23\textwidth]{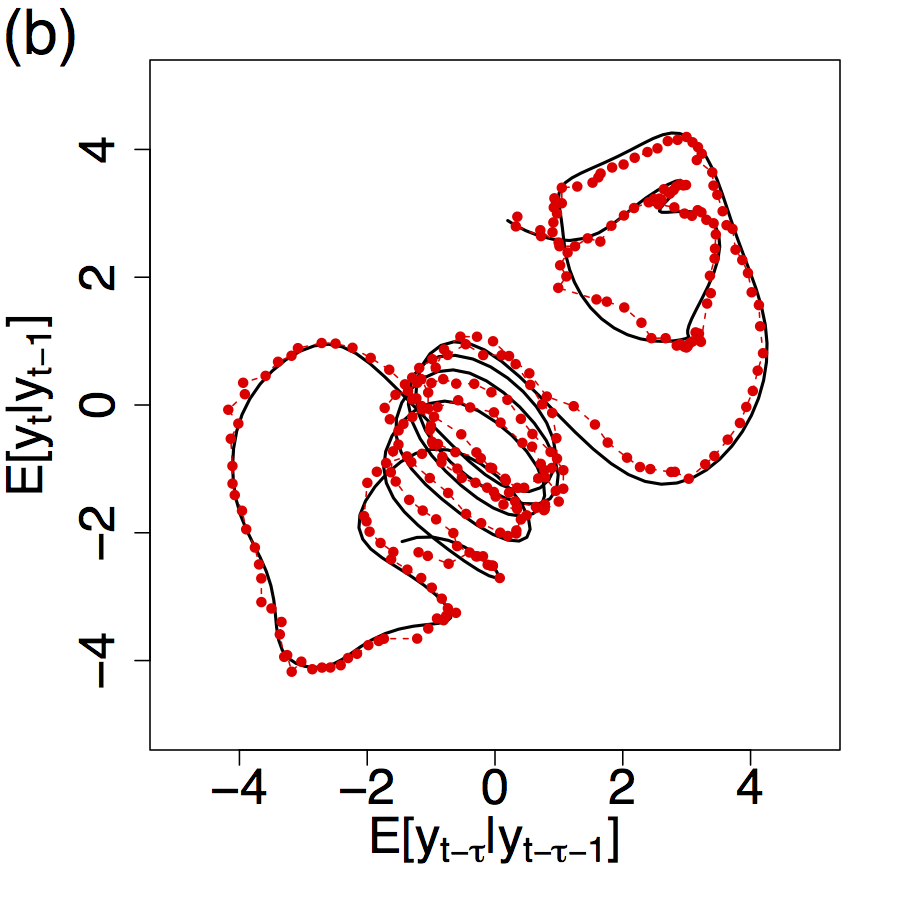}
  \caption{ (a) and (b) show the trajectories in a delay-time coordinate. The solid line is the ground truth and the circles ({\color{red}$\bullet$}) in (a) are the noisy observations and in (b) are the LSTM predictions. }\label{fig:Ikeda_pred}
\end{figure}

Next, the LSTM is tested against the Ikeda equation \cite{Ikeda80,Sprott},
\begin{equation}\label{eqn:Ikeda}
\frac{d y }{dt} = -y(t) + \alpha \sin ( y(t-\tau) ).
\end{equation}
The parameters are $\alpha = 6$ and $\tau = 1$, for which the Ikeda equation becomes chaotic\cite{Sprott}. Equation (\ref{eqn:Ikeda}) is solved by a third-order Adam-Bashforth method with the time step size of 0.001. The sampling interval is $\delta t = 0.05$. All other parameters, such as the noise level, $\Delta_\alpha$, and $\lambda$ are kept the same with the Mackey-Glass equation.

The phase portraits of $\yhat$ and the LSTM prediction are shown in figure \ref{fig:Ikeda_pred}. The attractor of the Ikeda equation has a more complex structure than the Mackey-Glass equation. Still, it is shown that the LSTM can reliably reconstruct the trajectory in the phase space from the noisy data. The prediction error for the Ikeda system is about 30\% of the noise,  $\langle [E[\yhat_{t+1}|\yhat_t] - y(t+\delta t)]^2 \rangle^{1/2} = 0.302 sd[\epsilon]$.

\begin{figure}
  \includegraphics[width=0.49\textwidth]{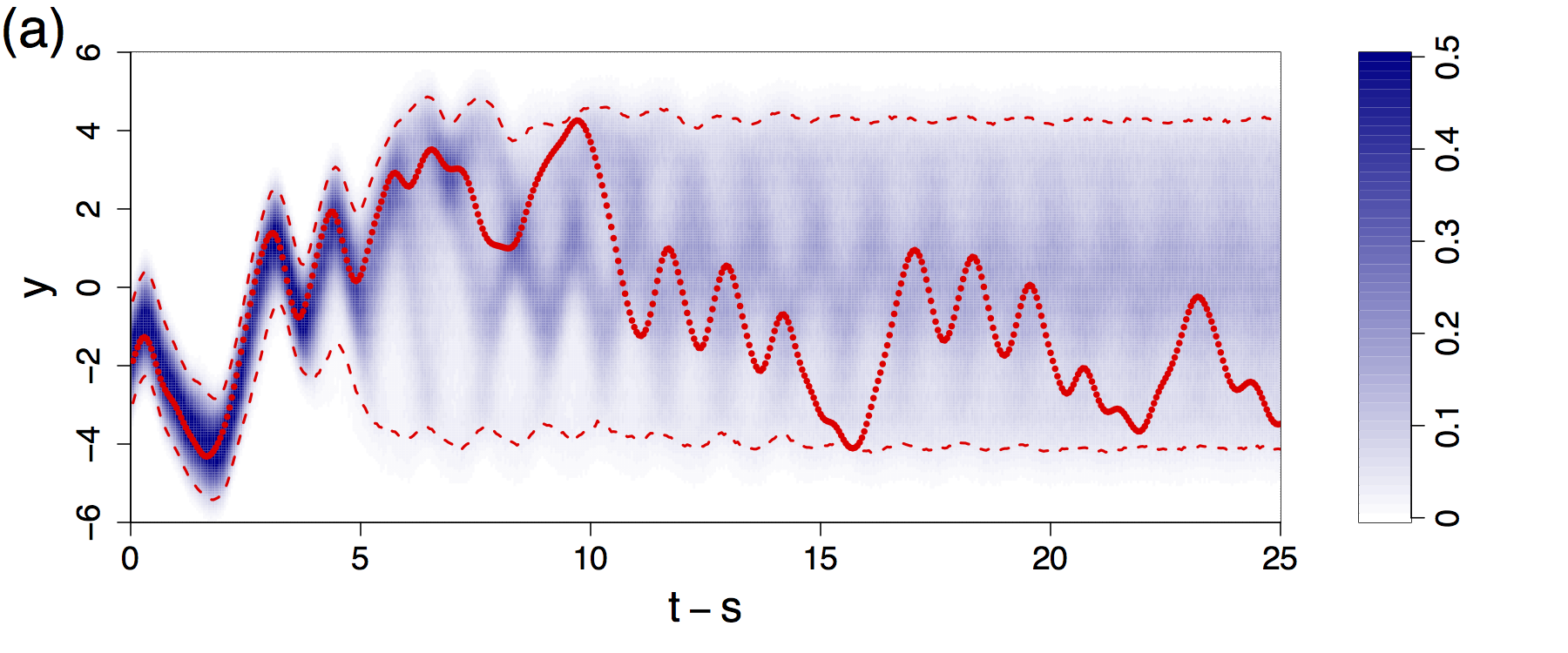}\\
  \includegraphics[width=0.49\textwidth]{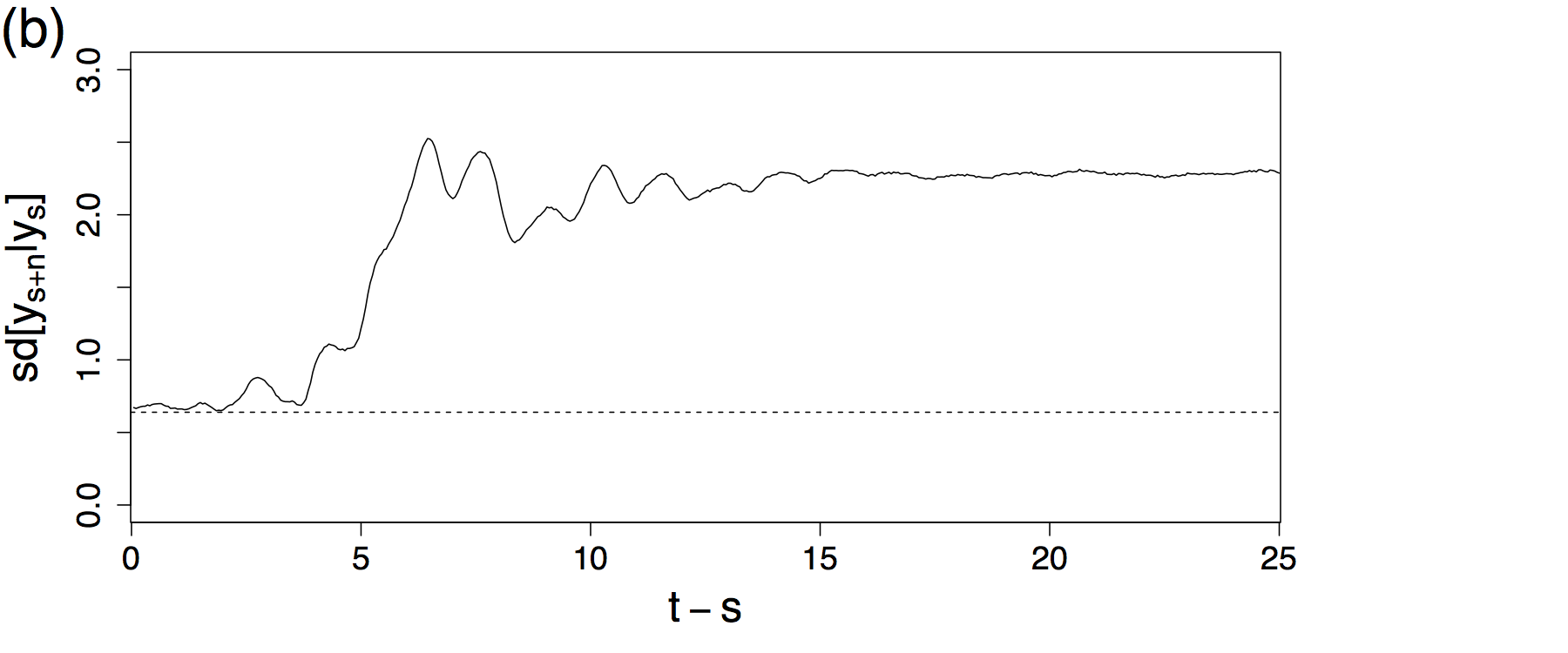}
    \caption{ (a) A 500-step forecast of the Ikeda equation. The contours represent the probability distribution conditioned on the last observation; $p(\yhat_{s+n}|\yhat_s)$. The dashed lines indicate the 95\% confidence interval and the solid circles ({\color{red}$\bullet$}) are the ground truth. (b) The predicted standard deviation in time. The dashed line indicates the noise level.}\label{fig:Ikeda_dist}
\end{figure}

Figure \ref{fig:Ikeda_dist} shows a 500-step forecast of the Ikeda equation with $N_s = 2\times10^4$. It is shown that up to $t=6$, or $120 \delta t$, $y(t)$ follows the high probability region of the LSTM. Unlike the Mackey-Glass equation, the confidence interval increases dramatically for $t > 5$ and then saturates for $t>10$. The confidence interval covers almost the range of $\yhat$, implying a long time forecast is impossible. The predicted STD in figure \ref{fig:Ikeda_dist} (b) clearly shows that the prediction uncertainty remains as the noise level for $t<3$, then starts to grow rapidly. Eventually, STD reaches a plateau for $t > 10$.

In this study, a deep learning model is developed for a model-free forecast of a chaotic dynamical system from noisy observations. The deep learning model consists of the LSTM network, which models the multiscale dynamics, and a softmax layer to approximate the probability distribution of the noisy dynamical system. The LSTM is trained by minimizing a regularized cross-entropy. Interestingly, even though only the noisy observations are shown, the LSTM makes a good prediction of the ground truth, i.e. the noise-free dynamical system.
Note that the delay times are 17$\delta t$ and 20$\delta t$ for the Mackey-Glass and Ikeda equations, respectively. To make a good prediction, the LSTM should be able to memorize the state of the system for many $\delta t$ and to know when to use the information. 
In a multiple-step forecast, it is shown that the prediction uncertainty dynamically changes over time and the ground truth lies in the 95\% confidence interval for a long time, e.g., 500-$\delta t$ forecasts. The results suggest that deep learning can provide a very powerful tool in data-driven modeling of complex dynamical systems.

\bibliography{LSTM_ref}

\end{document}